%% file: acl_latex.tex
\PassOptionsToPackage{hyphens}{url}
\documentclass[]{youtu}

\usepackage{times}
\usepackage{latexsym}
\usepackage{mathpazo}

\usepackage[T1]{fontenc}

\usepackage[utf8]{inputenc}

\usepackage{microtype}

\IfFileExists{inconsolata.sty}{\usepackage{inconsolata}}{}

\usepackage{graphicx}
\usepackage{wrapfig}
\usepackage{needspace}
\usepackage{float}
\usepackage{subcaption}
\usepackage{amsmath}
\usepackage{booktabs}
\usepackage{multirow}
\usepackage{colortbl}
\usepackage[most]{tcolorbox}
\usepackage{xspace}
\newcommand{\benchmark}{\emph{ElephantBench}\xspace}
\newcommand{\benchmarkicon}[1][1.15em]{\raisebox{-0.18\height}{\includegraphics[height=#1]{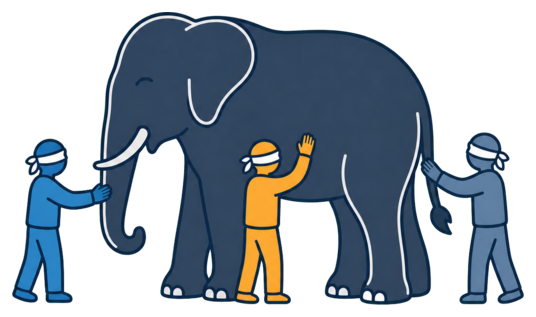}}}

\title{\benchmarkicon[1.35em]\hspace{0.3em}Blind Men and the Elephant: Probing the Epistemic Myopia of LLMs under Long-Tail Divergent Knowledge}

\author{
  Zhuoshi Pan\textsuperscript{1,2 $\clubsuit$},
  Junru Lu\textsuperscript{2 $\clubsuit$},
  Yan Qian\textsuperscript{3 $\clubsuit$},
  H. Vicky Zhao\textsuperscript{1$~\heartsuit$},
  Di Yin\textsuperscript{2},
  Xing Sun\textsuperscript{2$~\heartsuit$}
}

\affiliation{\textsuperscript{1}Tsinghua University\quad\textsuperscript{2}Tencent Youtu Lab\quad\textsuperscript{3}University of Warwick}

\project{https://tencent.github.io/ElephantBench}
\sourcecode{https://github.com/Tencent/ElephantBench}
\data{https://huggingface.co/datasets/Tencent/ElephantBench}

\correspondence{$\clubsuit$~Equal Contribution; $\heartsuit$~Corresponding Authors.}

\abstract{
Factual question answering (QA) typically assumes a single canonical answer, obscuring whether large language models (LLMs) retain divergent accounts of long-tail facts. To address this gap, we introduce \benchmark, a closed-book knowledge probe comprising 1,094 questions generated through an auditable graph-based pipeline. The pipeline retrieves related documents from a low-exposure web corpus, identifies naturally occurring disagreements, and converts them into multi-account QA records. Each answer is verified against the originating documents and authoritative public web sources and is then reviewed by human annotators. Across 32 models, even the strongest model recovers both accounts on only \textbf{52.4\%} of questions, while on nearly all remaining questions it recalls one account but omits the other. Scaling model size and inference-time reasoning improve recall but do not eliminate this incompleteness. Corpus analysis further shows that \textbf{\emph{exposure imbalance}} favors the dominant account, whereas greater minority-side exposure is associated with more complete recall. These findings establish \benchmark as a reproducible knowledge probe for diagnosing \textbf{\emph{epistemic myopia}} in parametric memory. More broadly, our graph-based benchmark construction pipeline provides an efficient and scalable way to turn long-tail corpora into source-traceable knowledge probes, supporting efforts to evaluate and advance the \textbf{\emph{epistemic rigour}} of next-generation LLMs.

}

\begin{document}
\maketitle

\begin{figure}[t]
\centering
\begin{tcolorbox}[
  colback=gray!7,
  colframe=gray!55!black,
  title={Source Evidence and a Matched Question Pair},
  fonttitle=\small\bfseries,
  fontupper=\footnotesize,
  fontlower=\footnotesize,
  boxrule=0.6pt,
  arc=1mm,
  boxsep=0.8mm,
  left=1mm,right=1mm,top=0.8mm,bottom=0.8mm
]
\textbf{Source A (IMDb)}\quad
\emph{``\ldots Mother Teresa was born on \textbf{August 26, 1910} \ldots as Agnes Gonxha Bojaxhiu. \ldots''}
\par\smallskip\hrule\smallskip
\textbf{Source B (Poem of Quotes)}\quad
\emph{``\ldots Mother Teresa, Agnes Gonxha Bojaxhiu, was born in Skopje, Macedonia on \textbf{August 27th, 1910}. \ldots''}
\par\smallskip\hrule\smallskip
\textbf{Named-entity question}\quad
What birth date was reported for Mother Teresa?
\par\smallskip\hrule\smallskip
\textbf{Clue-based question}\quad
What birth date was reported for the Albanian-born Roman Catholic nun who founded the Missionaries of Charity and received the 1979 Nobel Peace Prize?
\par\smallskip\hrule\smallskip
\textbf{Answers:}\
\textcircled{\scriptsize 1}\;\textbf{August 26, 1910}\
\textcircled{\scriptsize 2}\;\textbf{August 27, 1910}
\end{tcolorbox}
\vspace{-2mm}
\caption{Source evidence exhibiting factual disagreement and the corresponding matched question pair (named-entity vs.\ clue-based). Both formulations share the same verified answer set. Source texts are \textbf{concealed} during evaluation.}
\label{fig:paired-example-en}
\end{figure}

\section{Introduction}

Frontier large language models (LLMs) increasingly converge on canonical answers to common factual questions, making conventional question answering (QA) less discriminative \citep{xu2026deepseek,team2026kimi,lu2025youtu,team2026gemma}. To address this, we turn to the knowledge tail, where low-exposure facts offer a harder test of parametric memory. To mine such facts at scale, we invert the usual data-selection process and search the filtered remainder, $D_{\mathrm{low}}$, for challenging candidates \citep{li2024datacomplm,nguyen2025rewire,yu2025repro}. In this setting, relevant evidence is often distributed across multiple documents, and heterogeneous sources may report different accounts of the same underlying fact \citep{yuan-etal-2026-benchmarking}. 
A model may retain the dominant account while omitting less prevalent accounts \citep{lu2024scaling}, exhibiting the \textbf{\emph{epistemic myopia}} illustrated by the blind-men-and-the-elephant analogy (Figure~\ref{fig:paired-example-en}).
Therefore, our research question is: can an LLM recall a \textbf{\emph{long-tail fact}}, and how completely can it recover the \textbf{\emph{different verified accounts}} of that fact from its parametric memory?

Existing benchmarks address only parts of this problem. Long-tail QA tests whether models recall rare facts but typically assumes a single canonical answer \citep{mallen2023trust,kandpal2023longtail,zhu2025lpfqa}, leaving the completeness of memory untested. Knowledge-conflict benchmarks examine disagreement but usually adopt an open-book setting: they supply conflicting passages at inference time and test models' multi-hop reasoning over the provided evidence \citep{wang2024resolving,su2024conflictbank,pham2024whoqa}. Neither directly provides a closed-book test of whether LLMs' \textit{\textbf{parametric memory}} preserves different verified accounts of a long-tail fact.
We thus introduce \benchmark, a closed-book benchmark of 1,094 questions mined from a
low-exposure web corpus. Each item is \textit{\textbf{traceable to its source documents, externally verified, and human-reviewed}}. Without source documents, retrieval, or external tools, models must recover the different factual accounts from their parametric memory.
Our contributions are:
{\setlength{\leftmargini}{1em}
\begin{itemize}
    \setlength{\topsep}{0pt}
    \setlength{\partopsep}{0pt}
    \setlength{\itemsep}{0pt}
    \setlength{\parsep}{0pt}
    \item \textbf{A diagnostic probe of parametric memory.} Rather than collapsing model performance into a single accuracy score, our metrics separately diagnose whether a model remembers low-exposure facts and whether it recalls all verified accounts of those facts.
    \item \textbf{An efficient and reusable knowledge probe construction pipeline.} Our two-stage graph-based pipeline uses knowledge-point clustering and entity extraction to narrow the candidate space, and then employs LLMs to identify support and conflict edges. This design reduces the cost of graph construction and provides a reusable way to transform any low-exposure corpus into a knowledge probe for assessing LLMs' memory of long-tail facts.
    \item \textbf{A corpus-grounded analysis of memory completeness.} By counting the documents supporting each side of a disagreement, we reveal a clear exposure asymmetry: exposure to the more prevalent account helps models recall the fact, whereas exposure to its less prevalent counterpart is more closely associated with the completeness of that memory. This connects a measurable property of the corpus distribution to the memory failure mode, allowing the \textbf{\textit{probe}} to inform data curation rather than merely rank models.
\end{itemize}
}

\section{Related Work}

\paragraph{Long-tail knowledge evaluation and filtered-data recycling.}
Long-tail benchmarks define the tail using entity popularity \citep{mallen2023trust} or corpus frequency \citep{kandpal2023longtail}. LPFQA \citep{zhu2025lpfqa} and MINTQA \citep{he-etal-2026-mintqa} broaden evaluation to professional forums, recent knowledge, and less popular facts. However, these benchmarks generally score a single canonical answer, testing whether a rare fact can be recalled rather than whether models completely recall its different verified accounts. Separately, WRAP \citep{maini2024rephrasing} rewrites raw documents, while ReWire \citep{nguyen2025rewire}, RePro \citep{yu2025repro}, and SynPro \citep{yu2026synpro} recycle filtered documents for pretraining. 
WRAP++ \citep{zhou2026wrappp} incorporates cross-document relations, but still targets training-data synthesis. In contrast, \benchmark mines natural cross-source disagreements from the filtered remainder and turns them into a closed-book probe of whether different accounts coexist in parametric memory.

\paragraph{Knowledge conflicts in parametric memory and retrieved evidence.}
Knowledge-conflict benchmarks provide evidence at inference time to study conflicts involving parametric knowledge, retrieved evidence, misinformation, or temporal change \citep{chen-etal-2022-rich,wang2024resolving,su2024conflictbank,xu-etal-2024-knowledge-conflicts}. WikiContradict, ConfRAG, and WhoQA evaluate evidence-conditioned detection or reconciliation \citep{hou2024wikicontradict,yuan-etal-2026-benchmarking,pham2024whoqa}, while DYNAMICQA derives disputable facts from Wikipedia edit histories to probe intra-memory conflict and contextual adaptation \citep{marjanovic-etal-2024-dynamicqa}.
While these benchmarks use an open-book setting and
primarily test reading comprehension or multi-hop reasoning over the provided context, \benchmark withholds the source documents, requiring models to recover different accounts from parametric memory.

\paragraph{Multi-answer QA and source disagreement.}
AmbigQA enumerates plausible interpretations of underspecified questions \citep{min-etal-2020-ambigqa}, SituatedQA conditions answers on contextual variables such as time or location \citep{zhang-choi-2021-situatedqa}, and NATCONFQA evaluates disagreement through answer enumeration and pairwise agreement classification \citep{nachshoni-etal-2025-consensus}. These tasks center on query ambiguity, contextual dependence, or relations among candidate answers. In \benchmark, all accounts concern the \emph{same subject--attribute pair}, trace to naturally occurring web sources, and are withheld at inference time. This design tests whether parametric memory preserves differing verified accounts without imposing one account as the sole ground truth.

\begin{figure*}[t]
    \centering
    \includegraphics[width=\textwidth]{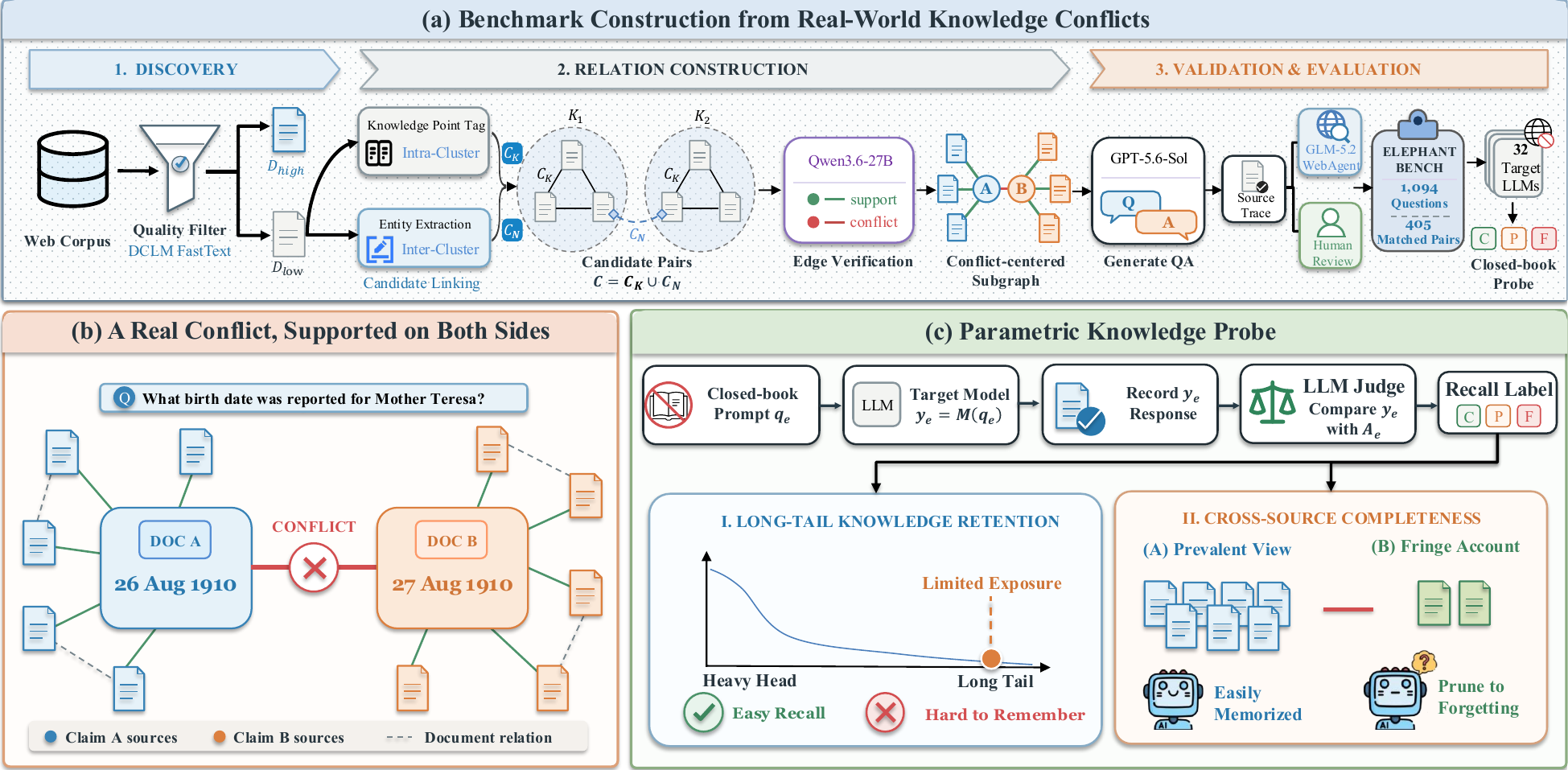}
    \caption{Overview of \benchmark as a systematic probe of parametric memory in LLMs. (a) Verified conflicts from low-exposure web documents are converted into 1,094 closed-book QA pairs. (b) Each benchmark item contains two independently supported accounts, testing whether models recall both. (c) The probe diagnoses \emph{epistemic myopia} through rare-fact retention and recall completeness under imbalanced cross-source exposure.}
    \label{fig:benchmark-overview}
\end{figure*}

\section{Tracing the Elephant}

Figure~\ref{fig:benchmark-overview} illustrates how \benchmark turns naturally occurring source disagreements into a controlled closed-book probe. The design separates two questions: whether a model can recall a low-exposure fact at all and, if so, whether it can recover all verified accounts reported for that fact.

\subsection{Task Formulation}

Let $D_{\mathrm{all}}$ denote a web corpus. Given a quality classifier $Q$ such as DCLM fastText \citep{li2024datacomplm} and a threshold $\tau$, we partition the corpus as
\begin{equation}
\begin{aligned}
D_{\mathrm{high}} &= \{z \in D_{\mathrm{all}} : Q(z) \geq \tau\}, \\
D_{\mathrm{low}}  &= \{z \in D_{\mathrm{all}} : Q(z) < \tau\}.
\end{aligned}
\end{equation}
$D_{\mathrm{high}}$ contains the documents retained by standard quality filtering, whereas $D_{\mathrm{low}}$ contains those removed by the filter.
By using $D_{\mathrm{low}}$, we retain documents containing rare facts that standard quality filtering removes, making \benchmark a sensitive probe of parametric memory.
Focusing on facts supported by only a handful of web documents makes low-exposure memory failures easier to detect, because such facts are less likely to receive additional exposure through targeted upsampling during pretraining.

Going beyond whether models retain low-exposure knowledge at all, we deliberately seek cross-document cases in which independently verified sources report incompatible accounts. These cases allow the probe to test whether parametric memory preserves all verified accounts represented in an item rather than only one.
We therefore represent $D_{\mathrm{low}}$ as a document graph: support edges connect documents concerning the same fact, whereas conflict edges identify incompatible accounts for the same subject--attribute pair. 
Verified conflict edges are then used as seeds for subgraph sampling and question-answer (QA) pair synthesis. Each resulting QA pair takes the form
\begin{equation}
x=(q,\mathcal{A}),\qquad \mathcal{A}=\{v_i\}_{i=1}^{k},
\end{equation}
where $q$ is a question about the target subject and $\mathcal{A}$ is the corresponding set of verified answers ($k\geq2$). Each $v_i$ must be traceable to a source retained for auditing but hidden from the evaluated model.

\subsection{Document Graph Construction}

To construct this graph at scale, we represent each document in $D_{\mathrm{low}}$ as a node:
\begin{equation}
G_D=(V_D,E_{\mathrm{sup}},E_{\mathrm{conf}}),
\end{equation}
where $V_D$ is the node set, $|V_D|=|D_{\mathrm{low}}|$, and $T:V_D\rightarrow D_{\mathrm{low}}$ maps each node to its full document. Support edges $E_{\mathrm{sup}}$ connect documents about the same specific fact, while conflict edges $E_{\mathrm{conf}}$ identify incompatible accounts for the same subject--attribute pair. The graph serves as an offline construction and audit structure: it brings dispersed evidence together, isolates disagreements for question-answer generation, and is never shown to the evaluated model.

\paragraph{Retrieving candidate pairs.}
A naive approach would examine $O(n^2)$ document pairs using an LLM judge, where $n=|V_D|$. We instead use lightweight knowledge-point and entity-based retrieval to identify candidate pairs and verify relations only between those candidates.
First, an LLM assigns each node $d$ a knowledge-point label $K(d)$ from its text $T(d)$, following the SuperGPQA taxonomy \citep{du2025supergpqa}. We then pair nodes assigned to the same label:
\begin{equation}
\mathcal{C}_{K}=\bigcup_k
\{\{d_i,d_j\}:d_i,d_j\in B_k,\ i<j\},
\end{equation}
where $B_k=\{d\in V_D:K(d)=k\}$ is the document set for knowledge cluster $k$. This yields $\sum_k\binom{|B_k|}{2}$ within-cluster comparisons, rather than exhaustively examining all $\binom{|V_D|}{2}$ document pairs.

However, related documents may be assigned different knowledge-point labels and thus fall into different clusters. To address this, we use named-entity recognition (NER) to retrieve cross-cluster pairs that share a normalized entity or event mention. Let $N(d)$ denote the set of normalized entity and event mentions extracted from $T(d)$. We implement this retrieval with an inverted index:
\begin{equation}
\mathcal{C}_{N}=\{\{d_i,d_j\}:N(d_i)\cap N(d_j)\neq\emptyset,\;K(d_i)\neq K(d_j)\}.
\end{equation}
Both mechanisms generate candidate pairs rather than directly determining graph edges. Their union, $\mathcal{C}=\mathcal{C}_{K}\cup\mathcal{C}_{N}$, limits relation classification to pairs sharing either a knowledge-point label or a normalized entity mention, thereby avoiding exhaustive comparison of all document pairs.

\paragraph{Inducing graph edges.}
Let $f_\zeta$ denote an LLM-based edge classifier. Given the full texts of a candidate pair $\{d_i,d_j\}\in\mathcal{C}$ and a single classification prompt $P$, it predicts a relation label $r_{ij}$:
\begin{equation}
r_{ij}=f_\zeta\!\left(T(d_i),T(d_j),P\right),
\end{equation}
where $r_{ij}\in\{\mathtt{none},\mathtt{support},\mathtt{conflict}\}$. The prompt $P$ provides instructions and demonstrations for all three labels. The \texttt{none} label indicates that the documents do not discuss the same specific fact. \texttt{support} indicates that they discuss the same subject--attribute pair and their reported accounts agree, whereas \texttt{conflict} indicates that they discuss the same pair but report incompatible accounts. Pairs labeled \texttt{support} and \texttt{conflict} form the support edge set $E_{\mathrm{sup}}$ and conflict edge set $E_{\mathrm{conf}}$, respectively. Because each candidate pair is classified once, graph induction only requires $|\mathcal{C}|$ edge-classifier calls.

\subsection{Question Generation and Filtering}
\label{sec:validation-en}

\paragraph{Synthesizing QA records.}
As illustrated in Figure~\ref{fig:benchmark-overview}, for each conflict edge $e\in E_{\mathrm{conf}}$, we extract a conflict-centered local subgraph $H_e=(V_e,E_e)$ containing the two endpoints and their support neighbors. The resulting subgraph brings together both incompatible accounts and the documents supporting each side. An LLM receives the graph structure and the full text $T(d)$ for every node $d\in V_e$, and generates a QA record consisting of a question $q_e$ and a reference-answer set $\mathcal{A}_e$:
\begin{equation}
\left(q_e,\mathcal{A}_e\right)
=f_\theta\!\left(e,H_e,\{T(d)\}_{d\in V_e}\right).
\end{equation}
The generated question $q_e$ queries the attribute on which the endpoints of $e$ disagree. The accompanying reference set $\mathcal{A}_e=\{v_i\}_{i=1}^{k}$ lists every distinct verified answer, each grounded in an evidence span from a document in the subgraph, as illustrated in Figure~\ref{fig:paired-example-en}. We generate two matched forms of $q_e$: a \emph{named-entity} question that states the target explicitly and a \emph{clue-based} counterpart that identifies it using at least two related attributes. Both forms share $\mathcal{A}_e$, allowing their score difference to isolate the effect of indirect identification.

\paragraph{Post-generation validation.}
We validate every synthesized record in three stages. First, an LLM reads the full text of all documents $d\in V_e$ and verifies that every $v_i\in\mathcal{A}_e$ is explicitly supported by at least one of them. Second, a web agent equipped with \texttt{WebSearch} and \texttt{WebFetch} independently searches for public evidence supporting each answer in every candidate QA pair. A record is retained only if both sides are supported by authoritative sources independent of the seed documents, such as Wikipedia. Third, human reviewers inspect both the original documents and the retrieved evidence, rejecting entity mismatches, unsupported answers, answer leakage, and question--fact mismatches. Records that fail any stage are discarded.

\subsection{Closed-Book Evaluation and Scoring}
\label{sec:task-evaluation-en}

Given $x_e=(q_e,\mathcal{A}_e)$, the target model receives only $q_e$ and produces $y_e=M(q_e)$, with no sources, graph structure, reference answers or external tools provided. This closed-book design turns \benchmark into a probe of parametric knowledge: it tests whether models remember low-exposure facts and whether they can recover all verified accounts associated with those facts.

Because free-form responses may express the same factual content in different ways, we use an LLM-as-a-judge for semantic evaluation. Given $(q_e,\mathcal{A}_e,y_e)$, it assigns one of three outcomes: (1) \emph{complete recall} when $y_e$ contains all reference answers and no incorrect answer, (2) \emph{partial recall} when it contains at least one but not all reference answers, and (3) \emph{failed recall} when it contains no reference answer or an incorrect answer. Let $N_C$, $N_P$, and $N_F$ denote the corresponding counts among $N$ responses. We report
\begin{equation}
C=\frac{N_C}{N},\qquad P=\frac{N_P}{N},\qquad F=\frac{N_F}{N}.
\end{equation}
Thus, $C+P+F=1$. Higher $C$ indicates better performance, while $P$ distinguishes incomplete but error-free recall from $F$. To isolate completeness among responses that recall at least one verified answer, we report conditional completeness $K$. Higher values of $K$ indicate better performance:
\begin{equation}
K=\frac{N_C}{N_C+N_P}=\frac{C}{C+P}.
\end{equation}

\section{Experiments}

\subsection{\benchmark{} Curation}

We apply the DCLM fastText quality classifier \citep{li2024datacomplm} to the RePro organic data \citep{yu2025repro} and retain its lower-scoring partition, $D_{\mathrm{low}}$. Qwen3.6-27B \citep{qwen2026qwen36} assigns knowledge-point labels following the SuperGPQA taxonomy \citep{du2025supergpqa}. 
\begin{wrapfigure}[11]{r}{0.5\columnwidth}
\centering
\includegraphics[width=0.92\linewidth,trim=7pt 6pt 4pt 6pt,clip]{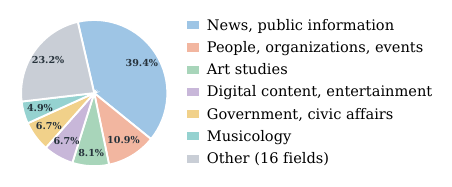}
\caption{Distribution of the 1,094 benchmark questions across knowledge fields. The six largest fields are shown separately, and the remaining 16 are grouped as Other.}
\label{fig:field-distribution-en}
\end{wrapfigure}
Documents sharing a label form within-cluster candidate pairs, while \mbox{T-NER}\footnote{\url{https://huggingface.co/tner/deberta-v3-large-ontonotes5}} \citep{ushio2021tner} retrieves additional cross-cluster pairs by matching shared entity and event mentions. Following the edge-verification prompt in Appendix Figure~\ref{fig:construction-prompts-en}, Qwen3.6-27B examines the full text of both documents in each candidate pair and assigns one of three labels: 
\texttt{support} for compatible accounts of the same fact, \texttt{conflict} for incompatible accounts of the same subject--attribute pair, and \texttt{none} otherwise. 
Pairs labeled \texttt{support} or \texttt{conflict} become the corresponding edges in the global document graph.

\input{tables/main_results_en}

We sample 4,127 conflict edges and expand each seed from its endpoints to their support neighbors, forming one local subgraph per edge. From each subgraph, GPT-5.6-Sol \citep{openai2026gpt56sol} generates two matched QA records: one in named-entity form and the other in clue-based form. Across the 4,127 subgraphs, this yields 8,254 candidate questions. We use GLM-5.2 \citep{glm5team2026glm5} as the model underlying the web-verification agent described in \S\ref{sec:validation-en}, and human reviewers conduct the final audit. After validation and deduplication, 1,094 QA pairs remain. Figure~\ref{fig:field-distribution-en} summarizes their distribution across 22 fields: \emph{News, public information} is the largest field (39.4\%), while the 16 fields outside the six largest collectively account for 23.2\%.

\subsection{Model Evaluation}

We evaluate 26 open-weight models and 6 proprietary systems. By default, reasoning is enabled. When reasoning effort is configurable, we set it to \textit{high}. We also compare against configurations with reasoning disabled, as analyzed in Section~\ref{sec:reasoning-effect-en}\footnote{When an interface does not allow reasoning to be disabled, we instead set reasoning effort to low.}. All models receive the closed-book prompt shown in Appendix Figure~\ref{fig:closed-book-prompt-en}(a). We use GPT-5.6-Sol \citep{openai2026gpt56sol} as the judge, applying the judge rubric shown in Figure~\ref{fig:closed-book-prompt-en}(b). Appendix~\ref{app:judge-reliability-en} evaluates its agreement with human annotations. For evaluation metrics, we report the \emph{complete recall} ($C$), \emph{partial recall} ($P$), \emph{failed recall} ($F$), and \emph{conditional completeness} ($K$) as introduced in Section~\ref{sec:task-evaluation-en}.

\subsection{Evaluation Results}

\paragraph{Even frontier models show a large gap between factual recall and completeness.}
Table~\ref{tab:main-results-en} reports two aspects of the knowledge probe: $C+P$ versus $F$ indicates whether a model recalls any verified answer to a long-tail fact, while $C$, $P$, and $K$ reveal whether that recall covers all of its documented accounts. The three strongest models almost always recall at least one verified answer, with failed recall rates ranging from only 2.19\% to 2.65\%, indicating that complete failure to recall the fact is rare. Complete recall, however, remains near 50\%: Kimi-K3 achieves the highest complete recall rate at 52.38\%, followed by Gemini-3.1-Pro at 50.37\% and GPT-5.5 at 50.18\%. Their partial recall rates remain similarly high, ranging from 45.25\% to 47.44\%. The dominant failure is therefore not failure to recall any verified answer, but incomplete recall across its documented accounts---an epistemic myopia that conventional single-answer QA can overlook.
\WFclear
\begin{wrapfigure}[14]{r}{0.45\columnwidth}
\centering
\vspace{-4mm}
\includegraphics[width=\linewidth]{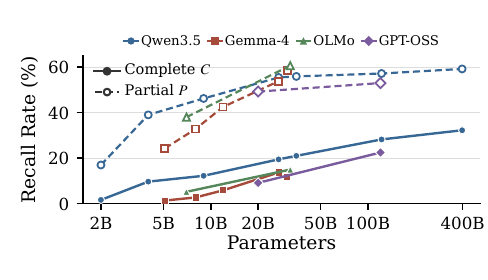}
\vspace{-8mm}
\caption{Within-family scaling trends for open-weight models. Solid and dashed lines show complete recall $C$ and partial recall $P$, respectively, on the same scale.}
\label{fig:model-scaling-en}
\vspace{-4mm}
\end{wrapfigure}
\paragraph{Effect of model scale.}
Complete recall improves substantially with model scale. Across open-weight models, the average complete recall rate is 5.48\% for models with fewer than 10B parameters, while the best model with more than 1T parameters reaches 52.38\%. The same trend holds within individual model families, as shown in Figure~\ref{fig:model-scaling-en}. For Qwen3.5, scaling from 2B to 397B raises $C$ from 1.65\% to 32.27\% and reduces $F$ from 81.35\% to 8.50\%, enabling the model to recall at least one verified answer on far more questions. However, the partial recall rate simultaneously rises from 17.00\% to 59.23\%, showing that much of this improved recall remains incomplete.
\WFclear
\begin{wrapfigure}[28]{r}{0.45\columnwidth}
\centering
\vspace{-4mm}
\includegraphics[width=\linewidth]{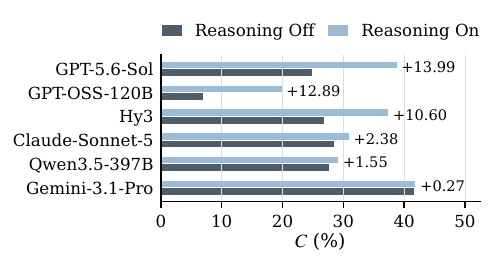}
\vspace{-8mm}
\caption{Effect of reasoning on complete recall. Dark and light bars show reasoning disabled and enabled, respectively. Labels report $\Delta C$ in percentage points.}
\label{fig:strong-reasoning-gains-en}
\vspace{6mm}
\includegraphics[width=\linewidth]{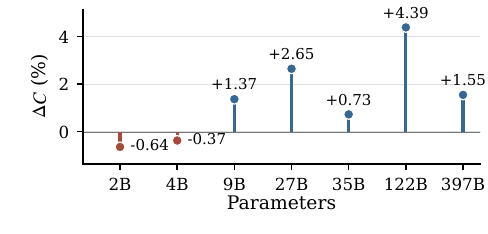}
\vspace{-8mm}
\caption{Effect of reasoning on complete recall across Qwen3.5 model sizes. Values report $\Delta C$ in percentage points, where positive values indicate higher $C$ with reasoning enabled.}
\label{fig:qwen-reasoning-effect-en}
\end{wrapfigure}
\paragraph{Effect of reasoning mode.}\label{sec:reasoning-effect-en}
Additional reasoning can substantially improve complete recall for some frontier models. Figure~\ref{fig:strong-reasoning-gains-en} shows representative gains: enabling reasoning increases the complete recall rate by 13.99 and 12.89 percentage points for GPT-5.6-Sol and GPT-OSS-120B, respectively. However, Figure~\ref{fig:qwen-reasoning-effect-en} reveals a scale-dependent pattern for Qwen3.5: enabling reasoning increases the complete recall rate by 0.73--4.39 percentage points from 9B to 397B, but reduces it by 0.64 and 0.37 percentage points at 2B and 4B, respectively. The two concrete regressions in Figure~\ref{fig:small-thinking-cases-en} (Appendix) illustrate cases in which a reasoning-enabled model treats one account as the consensus and omits the other, turning complete recall into partial recall. Reasoning therefore does not consistently improve recall completeness. For smaller models, additional deliberation can suppress rather than surface the less salient account.

\paragraph{Open-weight and proprietary models.}
Proprietary models occupy most of the leading positions, but the strongest overall result comes from an open-weight model. Kimi-K3 achieves a complete recall rate of 52.38\%, narrowly outperforming Gemini-3.1-Pro (50.37\%) and GPT-5.5 (50.18\%). Overall, two of the three best-performing systems are proprietary, but the open-weight Kimi-K3 ranks first.

\Needspace{0.18\textheight}
\section{Analysis}
\subsection{Factors Affecting Model Performance}

Because model-specific pretraining corpora are generally undisclosed, we approximate the relative exposure of competing accounts from our low-exposure web corpus. Pretraining data-processing pipelines often upsample or otherwise reweight curated high-quality and domain-specific data \citep{li2024datacomplm,xie2023doremi}, whereas low-exposure web documents may receive less targeted reweighting. We therefore use relative account frequencies in this corpus as an observational proxy for analyzing associations between source exposure and model performance.

\Needspace{0.36\textheight}
\begin{wrapfigure}{r}{0.45\columnwidth}
\centering
\vspace{-4mm}
\includegraphics[width=\linewidth]{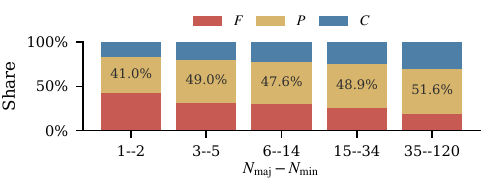}
\vspace{-1mm}
\includegraphics[width=\linewidth]{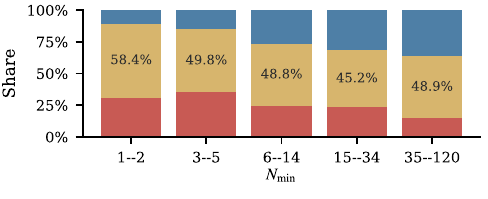}
\vspace{-8mm}
\caption{Recall outcomes by source exposure, binned by $N_{\mathrm{maj}}-N_{\mathrm{min}}$ (top) and $N_{\mathrm{min}}$ (bottom). Numbers in the yellow segments report partial recall $P$ (\%).}
\label{fig:frequency-performance-en}
\vspace{-2mm}
\end{wrapfigure}
Specifically, for each conflict, we count the documents supporting the two verified accounts, and denote the larger and smaller counts by $N_{\mathrm{maj}}$ and $N_{\mathrm{min}}$, respectively.
Figure~\ref{fig:frequency-performance-en} reveals two complementary patterns. As $N_{\mathrm{maj}}-N_{\mathrm{min}}$ grows, the supporting-document counts become more \textbf{imbalanced} and the partial recall rate generally \textbf{increases}, indicating that models more often recall only the \textbf{\textit{dominant account}}. Conversely, larger $N_{\mathrm{min}}$ is associated with higher $C$ and lower $P$, consistent with more complete recall when the \textbf{\textit{less prevalent account}} receives \textbf{greater} exposure.
We further disentangle the roles of exposure on the two sides using the joint regression reported in Appendix~\ref{app:joint-exposure-en} and Table~\ref{tab:exposure-regression-en}. The estimates reveal a clear asymmetry. A one-standard-deviation increase in exposure to the more frequently reported account is associated with a 14.18-percentage-point increase in $P$ and a 10.17-percentage-point decrease in $F$, suggesting that majority-side exposure primarily tracks whether models recall the fact at all. By contrast, the same increase for the less frequently reported account is associated with a 15.13-percentage-point increase in $C$ and a 15.41-percentage-point decrease in $P$, suggesting that minority-side exposure more closely tracks whether both verified accounts are recalled.

\subsection{Performance across Knowledge Domains}

\begin{wrapfigure}{r}{0.5\columnwidth}
\centering
\vspace{-4mm}
\includegraphics[width=\linewidth]{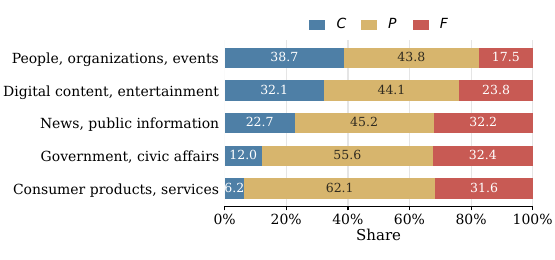}
\vspace{-8mm}
\caption{Outcome shares averaged over models for representative knowledge fields.}
\label{fig:field-performance-en}
\vspace{-2mm}
\end{wrapfigure}
Figure~\ref{fig:field-performance-en} shows a clear gradient across representative domains. Complete recall is highest for People, Organizations, and Events (38.7\%) and Digital Content and Entertainment (32.1\%), falls to 22.7\% for News Events and Public Information, and is lowest for Government and Civic Affairs (12.0\%) and Consumer Products and Services (6.2\%). In the two lowest-performing domains, partial recall dominates at 55.6\% and 62.1\%, respectively, indicating that models often recover one reported answer but not both. Qualitative inspection in Appendix~\ref{app:field-cases-en} reveals the same contrast. This field gradient may partly reflect the source composition of modern pretraining data mixtures. Wikipedia is often used as a source and repeatedly sampled \citep{touvron2023llama,soldaini-etal-2024-dolma}, which may give prominent people, organizations, and events greater exposure. In contrast, product prices, vote counts, and fine-grained dates may receive less exposure.

\Needspace{0.18\textheight}
\subsection{Cross-Model Oracle Coverage}

\begin{wrapfigure}{r}{0.45\columnwidth}
\centering
\vspace{-4mm}
\includegraphics[width=\linewidth]{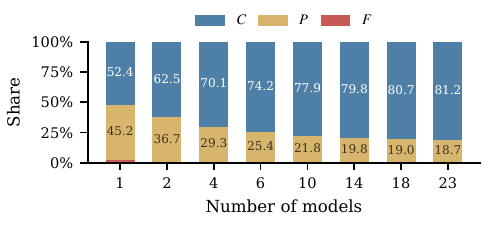}
\vspace{-8mm}
\caption{Outcome shares when models are greedily added to the oracle pool.}
\label{fig:oracle-curve-en}
\vspace{-2mm}
\end{wrapfigure}
Figure~\ref{fig:oracle-curve-en} shows the potential coverage of an oracle that selects the best response to each question. Starting with Kimi-K3, we greedily add the model that yields the largest increase in complete recall. As the model pool expands to all 32 configurations, $C$ rises from 52.4\% to 81.2\% and $F$ falls from 2.4\% to 0, with complete recall saturating after 23 configurations. This gain indicates substantial complementarity: collectively, the configurations recall at least one verified value for every question. Yet 18.8\% of questions remain partial. For these 206 questions, no configuration recovers all verified accounts. This incomplete recall indicates a limitation shared across models, particularly for less prevalent accounts.

\subsection{Robustness to Question Formulation}

We compare complete recall under two formulations of the same facts: named-entity questions that state the target explicitly and clue-based questions that identify it through descriptive clues. As shown in Appendix Figure~\ref{fig:formulation-robustness-en}, the mean difference across eight frontier models is only 0.65 percentage points (95\% CI $[-1.05, 2.31]$; $p=0.470$). Named-entity questions perform better for four models, and clue-based questions for the other four. We find no significant difference between the two formulations. These results suggest that performance is not driven primarily by question surface form.

\Needspace{0.30\textheight}
\subsection{PPL as an Efficient Evaluation Proxy}
\label{sec:ppl-proxy-en}

\begin{wrapfigure}{r}{0.45\columnwidth}
\centering
\vspace{-4mm}
\includegraphics[width=\linewidth]{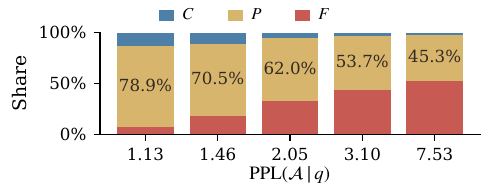}
\vspace{-8mm}
\caption{Recall outcomes across five perplexity bins for Qwen3.6-27B. Numbers in the yellow segments report partial recall $P$ (\%).}
\label{fig:ppl-binned-outcomes-en}
\vspace{-4mm}
\end{wrapfigure}
Generation-based evaluation requires autoregressive answer generation and evaluation by an LLM judge, making it computationally costly. We therefore examine conditional answer perplexity (PPL) as a lower-cost evaluation proxy. Appendix~\ref{app:ppl-probe-en} details the scoring procedure. Figure~\ref{fig:ppl-binned-outcomes-en} shows that, for the Qwen3.6-27B model, higher-PPL bins contain fewer complete responses and more failures. This close alignment with graded outcomes suggests that conditional PPL can serve as a computationally efficient proxy for generation-based evaluation.

\section{Conclusion}

\benchmark exposes the epistemic myopia hidden by single-answer factual QA: even the strongest model recalls all verified accounts for only slightly more than half of the questions. Complete recall is more closely associated with exposure to the less prevalent account than with the fact's overall frequency. Scaling model size and increasing inference-time reasoning effort improve recall of long-tail facts, but neither eliminates incomplete recall across verified accounts.
By combining long-tail facts with naturally occurring source disagreement in a closed-book setting, \benchmark distinguishes failure to recall a low-exposure fact from incomplete recall of its documented accounts.

\vspace{-4mm}
\section*{Limitations}

Our study has three main limitations. First, because evaluated models generally do not disclose their pretraining corpora, we rely on a publicly available pretraining corpus to construct the benchmark and use corpus-level frequencies as an observational proxy for factual exposure. Its distribution may differ from the data mixtures and sampling strategies used to train any particular model. Therefore, our exposure-related findings should be interpreted as associations rather than direct measurements of model-specific training data. Second, although the benchmark contains 1,094 questions spanning 22 knowledge fields, it cannot exhaustively represent the breadth of long-tail knowledge or the diverse ways in which sources may disagree. Future work could expand its scale and domain coverage while incorporating a broader range of disagreement types. Third, we evaluate 32 representative models across open-weight and proprietary families. Our evaluation therefore does not cover models released after this study, every available model size, or every reasoning configuration.

\section*{Ethics Statement}

Each retained account is independently supported by authoritative public web evidence, but may still reflect source error or misinformation. Human review removes unnecessary personal information, sensitive attributes, harmful allegations, and objectionable content. Appendices~\ref{app:artifact-release-en} and~\ref{app:human-annotation-en} document the release and screening procedures. \benchmark is intended only for factual-knowledge research and model evaluation, not surveillance, profiling, harassment, or other harmful uses. AI assistants were used for language editing and code debugging. The authors verified all assisted material and remain responsible for the final work.

\begingroup
\sloppy
\setlength{\bibsep}{0pt}
\bibliography{custom,references}
\endgroup

\clearpage
\appendix

\section{Prompts in Benchmark Curation}
\label{app:prompts-en}

The prompts used throughout benchmark construction are summarized in Figure~\ref{fig:construction-prompts-en}. They cover knowledge-point labeling, support and conflict edge induction, QA synthesis from conflict-centered subgraphs, and external verification of each reported value. At every stage, the model receives the relevant full documents and returns structured outputs for the next stage.

\begin{figure*}[t]
\centering\small
\begin{tcolorbox}[
  colback=gray!7,
  colframe=gray!55!black,
  title={Core Instructions for Evaluation-Data Construction},
  fonttitle=\bfseries,
  boxrule=0.6pt,
  arc=1mm,
  boxsep=1mm,
  left=1.5mm,right=1.5mm,top=1mm,bottom=1mm
]
\textbf{Knowledge labeling}\quad
\emph{Read the full document and select exactly one label---the most specific applicable label in the supplied SuperGPQA discipline--field--subfield hierarchy. Return the three levels as JSON; do not invent labels or classify from navigation and boilerplate alone.}
\par\medskip\hrule\medskip
\textbf{Support and conflict judgments}\quad
\emph{Compare both full documents. Mark support only when they concern the same concrete entity, event, method, or question. Mark conflict only when they assign values to the same attribute of the same subject that cannot both hold unconditionally. Return the relation, subject, attribute, values, and a verbatim evidence span for each side. A shared entity or broad topic is insufficient.}
\par\medskip\hrule\medskip
\textbf{Question generation}\quad
\emph{Given a candidate conflict edge, its local subgraph, and the full documents, generate a question that neither leaks an answer nor announces the conflict. List every distinct account supported by a verbatim span. Do not introduce outside knowledge. Return strict JSON with the question, values, evidence documents, and evidence spans.}
\par\medskip\hrule\medskip
\textbf{Agentic external verification}\quad
\emph{Search separately for every candidate value, prioritizing Wikipedia and reliable public sources. Verify that an opened page supports the exact subject--attribute--value claim; reject entity collisions, circular reposting, and support available only from the seed source. Return URLs, evidence excerpts, per-side support decisions, and a final keep/reject decision. Reject the instance if either side lacks independent support.}
\end{tcolorbox}
\caption{Core instructions used at each construction stage. Each prompt additionally receives the relevant documents, candidate labels, document pair, or local subgraph.}
\label{fig:construction-prompts-en}
\end{figure*}

\section{Candidate-Pair Reduction from Knowledge-Point Clustering}
\label{app:kp-clustering-efficiency-en}

We quantify the computational savings provided by knowledge-point clustering before pairwise relation judgment. For $n=|V_D|$ documents, exhaustive comparison would require
\begin{equation}
M_{\mathrm{all}}=\binom{n}{2}
\end{equation}
pairwise LLM calls. Grouping documents by their assigned knowledge point instead yields the within-cluster candidate set
\begin{equation}
M_{\mathrm{KP}}=\sum_k \binom{|B_k|}{2},
\end{equation}
where $B_k$ contains the documents assigned to knowledge point $k$. In our construction corpus,
\begin{equation}
\frac{M_{\mathrm{KP}}}{M_{\mathrm{all}}}\leq 6.6\%,
\qquad
\frac{M_{\mathrm{all}}}{M_{\mathrm{KP}}}\geq 15.2.
\end{equation}
Thus, knowledge-point clustering removes at least 93.4\% of the naive pair space and reduces the number of pairwise LLM judgments by at least \textbf{15.2$\times$}. This comparison isolates the computational gain achieved by the clustering step itself.

\section{Evaluation Settings}
\label{app:model-inference-en}

All target models receive the system instruction and user-message template shown in Figure~\ref{fig:closed-book-prompt-en}(a). No source document, reference answer, or scoring rubric is included. After the target-model response has been recorded, the GPT-5.6-Sol judge receives the question, verified answers, and model answer under the instruction shown in Figure~\ref{fig:closed-book-prompt-en}(b). The judge receives neither source documents nor external tools.
\begin{figure*}[t]
\centering
\begin{tcolorbox}[
  colback=gray!7,
  colframe=gray!55!black,
  title={(a) Closed-book system instruction},
  fonttitle=\bfseries,
  boxrule=0.6pt,
  arc=1mm,
  boxsep=1mm,
  left=1.5mm,right=1.5mm,top=1mm,bottom=1mm
]
\small\emph{You are answering a general-knowledge question using only your own training-time knowledge. You have no access to search, browsing, or any external tool, and no source documents have been given to you---rely solely on what you already know. If you recall that different reputable sources report different values for this fact, state all the values you recall and briefly note that sources disagree. If you only know one value, give only that value; if you genuinely do not know, say so rather than guessing. Keep the answer to at most three sentences.}
\par\smallskip\hrule\smallskip
\textbf{User message}\quad\texttt{<benchmark question>}
\end{tcolorbox}
\medskip
\begin{tcolorbox}[
  colback=gray!7,
  colframe=gray!55!black,
  title={(b) Judge prompt},
  fonttitle=\bfseries,
  boxrule=0.6pt,
  arc=1mm,
  boxsep=1mm,
  left=1.5mm,right=1.5mm,top=1mm,bottom=1mm
]
\footnotesize
\textbf{System instruction.}\quad Evaluate factual recall using only the supplied question, candidate golds, and model answer. Do not use external knowledge or tools, and judge semantic meaning rather than exact wording. Apply the following four steps in order:\par\smallskip
\textbf{1. Identify the requested slot.}\quad Determine what the question asks for and any relevant distinction, such as total versus subgroup, birth versus baptism, event versus publication, regional date, timeline stage, or counting convention.\par
\textbf{2. Assess every gold in input order.}\quad Count direct statements and clear explanatory coverage. Accept paraphrases, aliases, reasonable rounding or precision, conversions, categorical equivalents, and explanations that connect a gold to a source, definition, or timeline discrepancy. For a compound gold expressing alternatives or a range, require the proposition's meaning rather than every literal token. Do not count coincidental mentions or values for a different answer slot.\par
\textbf{3. Record only material contradictions.}\quad A contradiction is a non-gold value clearly endorsed for the requested slot that makes the answer wrong or materially ambiguous. Do not penalize contextual details merely because they are absent from the gold set, including explicitly rejected values, background facts, range bounds, reasonable approximations, subgroup/total breakdowns, regional versions, timeline stages, or additional source-reported variants used to explain uncertainty.\par
\textbf{4. Assign credit mechanically.}\quad Any material contradiction or zero covered golds $\Rightarrow$ \texttt{no\_credit}; otherwise, all golds covered $\Rightarrow$ \texttt{full\_credit}; otherwise $\Rightarrow$ \texttt{partial\_credit}. With one gold, coverage without a material contradiction receives \texttt{full\_credit}.\par\smallskip
\textbf{User message}\quad\texttt{<question; candidate gold values; model answer>}\par
\textbf{Scoring rubric.}
\begin{itemize}
\setlength{\itemsep}{0pt}\setlength{\topsep}{1pt}\setlength{\parsep}{0pt}\setlength{\parskip}{0pt}
\item \texttt{full\_credit}: every candidate gold is semantically covered, with no material contradiction.
\item \texttt{partial\_credit}: at least one but not every candidate gold is covered, with no material contradiction.
\item \texttt{no\_credit}: no candidate gold is covered, or a material contradiction is present.
\end{itemize}
\textbf{Required JSON output.}\quad \texttt{<gold\_assessments; material\_contradictions; credit; reasoning>}.
\end{tcolorbox}
\caption{Evaluation prompts. (a) Every target model receives the same closed-book instruction, with the original benchmark question replacing the placeholder. (b) After response generation, the judge receives the structured judging input and applies the full four-step rubric for semantic coverage.}
\label{fig:closed-book-prompt-en}
\end{figure*}

Main results follow each model's recommended generation recipe. API models, Nemotron, and Seed-OSS use temperature $0$. Qwen3.5/3.6 models use temperature $1.0$, top-$p=0.95$, top-$k=20$, min-$p=0$, and enable thinking. Gemma-4 models use temperature $1.0$, top-$p=0.95$, and top-$k=64$. OLMo models use temperature $0.6$ and top-$p=0.95$. GPT-OSS models use temperature $1.0$, top-$p=1.0$, and high reasoning effort. The output budget is 32,768 tokens. Throughout evaluation, no model has access to search, browsing, or retrieval tools, ensuring the same strictly closed-book setting for every model.

\Needspace{0.45\textheight}
\section{Reliability of the LLM Judge}
\label{app:judge-reliability-en}

\begin{wrapfigure}{r}{0.5\columnwidth}
\centering
\vspace{-4mm}
\includegraphics[width=\linewidth]{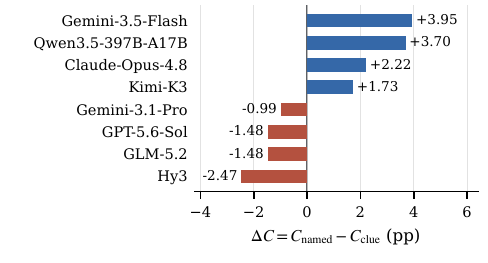}
\vspace{-8mm}
\caption{Complete recall differences between named-entity and clue-based questions for frontier models. Positive values favor named-entity questions.}
\label{fig:formulation-robustness-en}
\end{wrapfigure}
To assess the reliability of the GPT-5.6-Sol judge, three graduate annotators in computer science and artificial intelligence independently evaluated responses from four frontier models using the same three-way criteria. Table~\ref{tab:judge-human-agreement-en} reports 90.13--93.36\% exact agreement between human annotations and judge labels, with Cohen's $\kappa$ ranging from 0.815 to 0.877. The model ranking produced by the judge matches that obtained from the human annotations, and its complete recall rates differ from the human results by only 1.44 percentage points on average. These results show that the automatic evaluation closely matches human judgment and provides a reliable basis for our model comparisons.

\input{tables/judge_human_agreement_en}

\section{Robustness to Sampling Temperature}
\label{app:temperature-ablation-en}

We conduct a controlled ablation of the sampling temperature to test whether our conclusions are sensitive to this hyperparameter. For Qwen3.5-35B-A3B, Qwen3.6-35B-A3B, GPT-OSS-120B, and Gemma-4-31B, we compare outputs generated with $T{=}0$ and $T{=}1$ on all 1,094 questions while holding the checkpoint, prompt, reasoning configuration, output budget, and all other sampling parameters fixed. The same GPT-5.6-Sol judge scores every condition, and generation failures contribute to $F$ under a fixed denominator.
Table~\ref{tab:temperature-ablation-en} shows little sensitivity to the temperature setting: $\lvert\Delta C\rvert\leq0.46$ percentage points and $\lvert\Delta K\rvert\leq0.77$ percentage points across all four models. Exact paired McNemar tests detect no significant change in complete recall (all $p\geq0.53$). Every case-clustered bootstrap 95\% confidence interval for $\Delta C$ contains zero, and the model ranking remains unchanged. These results indicate that the main conclusions about recall are robust across the two temperature settings.

\input{tables/temperature_ablation_en}

\begin{figure*}[t]
\centering\small
\begin{tcolorbox}[colback=gray!5,colframe=gray!55!black,boxrule=0.6pt,arc=1mm,title={Case A: Qwen3.5-2B---Frederick Douglass's Birth Year},fonttitle=\bfseries]
\textbf{Question}\quad What birth year is reported for Frederick Douglass, the former slave who escaped in 1838 and published the \emph{Narrative of the Life of Frederick Douglass}?
\par\medskip\hrule\medskip
\textbf{Gold Answers}\quad \textcircled{\scriptsize 1}\;\textbf{1817}\qquad \textcircled{\scriptsize 2}\;\textbf{1818}
\par\medskip\hrule\medskip
\textbf{Non-Thinking---complete recall}\quad
\emph{``Frederick Douglass was born in 1818. While some sources suggest he may have been born in 1817, the most widely accepted and cited birth year is 1818.''}
\par\medskip\hrule\medskip
\textbf{Thinking---partial recall}\quad
\emph{``Frederick Douglass was reported to be born in 1818. Historical biographies and official records consistently confirm this date. While occasional variations exist in specific genealogical datasets, 1818 remains the generally accepted year.''}
\end{tcolorbox}
\vspace{0.35em}
\begin{tcolorbox}[colback=gray!5,colframe=gray!55!black,boxrule=0.6pt,arc=1mm,title={Case B: Qwen3.5-4B---Marilyn Monroe's Reported Death Date},fonttitle=\bfseries]
\textbf{Question}\quad What date is reported in various accounts for the death of Marilyn Monroe?
\par\medskip\hrule\medskip
\textbf{Gold Answers}\quad \textcircled{\scriptsize 1}\;\textbf{August 5, 1962}\qquad \textcircled{\scriptsize 2}\;\textbf{August 4, 1962}
\par\medskip\hrule\medskip
\textbf{Non-Thinking---complete recall}\quad
\emph{``Marilyn Monroe is most commonly reported to have died on August 5, 1962. However, some biographies and accounts suggest the date may be August 4 due to discrepancies regarding the time of her death.''}
\par\medskip\hrule\medskip
\textbf{Thinking---partial recall}\quad
\emph{``According to official records and widely accepted biographical accounts, Marilyn Monroe died on August 5, 1962. While some historical debates exist regarding the precise hour and circumstances of her passing, no reputable sources report a conflicting calendar date.''}
\end{tcolorbox}
\caption{Representative recall regressions observed after enabling reasoning.}
\label{fig:small-thinking-cases-en}
\end{figure*}

\Needspace{0.46\textheight}
\section{Performance across Conflict Mechanisms}
\label{app:conflict-mechanisms-en}

\begin{wrapfigure}{r}{0.5\columnwidth}
\centering
\vspace{-4mm}
\includegraphics[width=\linewidth]{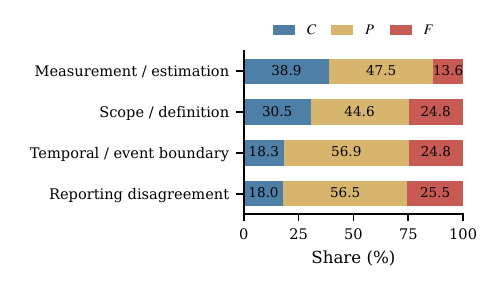}
\vspace{-8mm}
\caption{Recall outcomes across conflict mechanisms, averaged over 32 models. Numbers inside the bars report percentages. From top to bottom, the categories contain 86, 459, 291, and 258 questions derived from 30, 175, 103, and 94 underlying conflicts.}
\label{fig:conflict-mechanisms-en}
\end{wrapfigure}
To identify which kinds of disagreement are most difficult, we use GPT-5.6-Sol to classify each conflict into one of four mechanisms based on its paired source claims and external-verification record: (1) temporal changes or event boundaries, (2) scope or definition differences, (3) measurement or estimation conventions, and (4) direct reporting disagreements.
Figure~\ref{fig:conflict-mechanisms-en} shows substantial variation across mechanisms. Measurement-related conflicts are easiest, with $C=38.9\%$ and $F=13.6\%$, followed by scope or definition differences at $C=30.5\%$. Temporal and reporting disagreements are considerably harder: their $C$ values are only 18.3\% and 18.0\%, while $P$ reaches 56.9\% and 56.5\%, respectively. A case-level Kruskal--Wallis test confirms that complete recall differs across mechanisms ($H=49.14$, $p<10^{-9}$). Reasoning yields the largest average gain for temporal conflicts: across 22 matched reasoning-on/off comparisons, it raises $C$ by 4.12 percentage points on average, with positive gains for 15 pairs (Wilcoxon $p=0.011$). Mean gains for other mechanisms range from 1.27 to 1.94 percentage points and are not statistically significant.

\section{Joint Exposure Regression}
\label{app:joint-exposure-en}

For each item, we count the RePro documents that independently support each of its two verified accounts, excluding the provenance documents used to construct the item. For item $i$, $N_{\mathrm{maj},i}$ and $N_{\mathrm{min},i}$ denote the larger and smaller supporting-document counts, respectively. Let $C_i$, $P_i$, and $F_i$ be the proportions of evaluated model configurations producing each recall outcome on item $i$. Because the exposure counts are right-skewed, we log-transform and standardize them as
\begin{equation}
z_{s,i}=\frac{\log(1+N_{s,i})-\mu_s}{\sigma_s},
\qquad s\in\{\mathrm{maj},\mathrm{min}\},
\end{equation}
where $\mu_s$ and $\sigma_s$ are the mean and standard deviation of the log-transformed counts. We then fit the following item-level regression separately for $Y\in\{C,P,F\}$:
\begin{equation}
Y_i=\alpha+\beta_{\mathrm{maj}}z_{\mathrm{maj},i}
+\beta_{\mathrm{min}}z_{\mathrm{min},i}
+\epsilon_i.
\end{equation}
Each $\beta_s$ estimates the association with the exposure of one account while holding the other side fixed. Because exposure is standardized and outcomes are proportions, $100\beta_s$ is the corresponding percentage-point change for a one-standard-deviation increase. Importantly, these coefficients capture observational associations rather than causal effects.

\input{tables/exposure_regression_en}

\section{Cases from Shared Model Blind Spots}
\label{app:blind-cases-en}

Figure~\ref{fig:blind-cases-en} presents four representative questions for which none of the 32 evaluated models recovers all verified accounts. The cases cover a corrected vote tally, conflicting reports of a cause of death, differing interpretations of a prison sentence, and the regional scope of a release date. Each panel shows the paired source passages retained during benchmark verification, the corresponding question, and its two verified answers. We further cross-check the reported answers against independent public sources.\footnote{Case A: \href{https://www.parliament.nz/en-nz/pb/debates/debates/50HansD_20120829_00000032/}{official New Zealand Parliament record}. Case B: \href{https://doi.org/10.1136/bmj.39451.528576.DB}{contemporaneous \emph{BMJ} article}.} The audit confirms that every reference answer is supported by an authentic source and accurately reflects its report, yet no model recovers all supported accounts.

\begin{figure*}[t]
\centering\small
\begin{tcolorbox}[colback=gray!5,colframe=gray!55!black,boxrule=0.6pt,arc=1mm,title={Case A: First Reading of the Marriage Amendment Bill},fonttitle=\bfseries]
\textbf{Source A (Box Turtle Bulletin)}\quad
\emph{``\ldots In the first of three votes, the New Zealand parliament has overwhelmingly voted for marriage. Labour MP Louisa Wall's Marriage Amendment Bill was passed with \textbf{78 in favour and 40 against}. This margin suggests that there is a quite good chance that the bill will make it through all three readings to become law before the end of the year. \ldots''}\hfill{\scriptsize(Excerpt truncated for space.)}
\par\medskip\hrule\medskip
\textbf{Source B (Stuff)}\quad
\emph{``\ldots Well-known New Zealanders have banded together in a new video to show solidarity for the proposed Marriage Amendment Bill. The Definition of Marriage Amendment Bill, sponsored by Labour MP Louisa Wall, passed its first reading in Parliament in August by a convincing \textbf{80 votes to 40}. Video co-producer, television personality Tamati Coffey said filming the video has been history in the making. \ldots''}\hfill{\scriptsize(Excerpt truncated for space.)}
\par\medskip\hrule\medskip
\textbf{Question}\quad How many votes in favour were reported for the first reading of the Marriage (Definition of Marriage) Amendment Bill in New Zealand?
\par\medskip\hrule\medskip
\textbf{Gold Answers}\quad \textcircled{\scriptsize 1}\;\textbf{78}\qquad \textcircled{\scriptsize 2}\;\textbf{80}
\end{tcolorbox}
\vspace{0.35em}
\begin{tcolorbox}[colback=gray!5,colframe=gray!55!black,boxrule=0.6pt,arc=1mm,title={Case B: Reports of Benazir Bhutto's Cause of Death},fonttitle=\bfseries]
\textbf{Source A (Atlantic Free Press)}\quad
\emph{``\ldots The announcement, by Pakistani officials, that former Prime Minister Benazir Bhutto died from a \textbf{skull fracture} when falling against the wall of her sports utility vehicle, during the recent rally, and not from bullets from a gun that was aimed directly at her, or schrapnal [sic] from a suicide bomb \ldots''}\hfill{\scriptsize(Excerpt truncated for space.)}
\par\medskip\hrule\medskip
\textbf{Source B (contemporaneous CNN report)}\quad
\emph{``\ldots Pakistan's former Prime Minister Benazir Bhutto was assassinated Thursday after addressing a large gathering of her supporters. Bhutto died of a \textbf{gunshot wound to the neck}, the Pakistani Interior Ministry said. The attacker then blew himself up, and the bomb attack killed at least 22 others, doctors said. \ldots''}\hfill{\scriptsize(Excerpt truncated for space.)}
\par\medskip\hrule\medskip
\textbf{Question}\quad What cause of death did Pakistani officials initially announce for former Prime Minister Benazir Bhutto following her death during a political rally in Rawalpindi?
\par\medskip\hrule\medskip
\textbf{Gold Answers}\quad \textcircled{\scriptsize 1}\;\textbf{skull fracture}\qquad \textcircled{\scriptsize 2}\;\textbf{gunshot wound to the neck}
\end{tcolorbox}
\caption{Four verified examples of shared model blind spots. Each case contains two source-supported answers, but no evaluated model recovers both. Cases C and D continue on the next page.}
\label{fig:blind-cases-en}
\end{figure*}

\begin{figure*}[t]
\ContinuedFloat
\centering\small
\begin{tcolorbox}[colback=gray!5,colframe=gray!55!black,boxrule=0.6pt,arc=1mm,title={Case C: Reported Status of the Dolce \& Gabbana Prison Sentence},fonttitle=\bfseries]
\textbf{Source A (Styleite)}\quad
\emph{``\ldots Earlier this morning, an Italian court found the Dolce \& Gabbana designers guilty of tax evasion and \textbf{sentenced them each to one year and eight months of jail time}. As we would presume, the designers are expected to appeal the decision. \ldots''}\hfill{\scriptsize(Excerpt truncated for space.)}
\par\medskip\hrule\medskip
\textbf{Source B (InsideCounsel)}\quad
\emph{``\ldots An Italian court handed down a \textbf{20-month suspended prison sentence} to the designer duo. The court also ordered the designers to pay a fine following the tax-evasion ruling. \ldots''}\hfill{\scriptsize(Excerpt truncated for space.)}
\par\medskip\hrule\medskip
\textbf{Question}\quad What was the reported nature of the prison sentence handed to Domenico Dolce and Stefano Gabbana in the Italian court ruling on their tax-evasion case?
\par\medskip\hrule\medskip
\textbf{Gold Answers}\quad \textcircled{\scriptsize 1}\;\textbf{active jail sentence}\qquad \textcircled{\scriptsize 2}\;\textbf{suspended sentence}
\end{tcolorbox}
\vspace{0.35em}
\begin{tcolorbox}[colback=gray!5,colframe=gray!55!black,boxrule=0.6pt,arc=1mm,title={Case D: Reported European Release Date of Kingdom Hearts Re:coded},fonttitle=\bfseries]
\textbf{Source A (Siliconera)}\quad
\emph{``\ldots Square Enix just updated the Kingdom Hearts Re:coded site with the release date, \textbf{January 14, 2011 for Europe}. January 14th, 2011---that is when Square Enix are publishing Kingdom Hearts Re:coded in Europe. \ldots''}\hfill{\scriptsize(Excerpt truncated for space.)}
\par\medskip\hrule\medskip
\textbf{Source B (GameZone)}\quad
\emph{``\ldots Kingdom Hearts Re:coded, the Nintendo DS remake of coded, was released on October 7, 2010 in Japan and will be released on \textbf{January 11, 2011, in the west}. \ldots''}\hfill{\scriptsize(Excerpt truncated for space.)}
\par\medskip\hrule\medskip
\textbf{Question}\quad What date was reported for the release of Kingdom Hearts Re:coded in Europe?
\par\medskip\hrule\medskip
\textbf{Gold Answers}\quad \textcircled{\scriptsize 1}\;\textbf{January 14, 2011}\qquad \textcircled{\scriptsize 2}\;\textbf{January 11, 2011}
\end{tcolorbox}
\caption[]{Shared model blind-spot cases (continued).}
\end{figure*}

\section{Illustrative Cases across Knowledge Domains}
\label{app:field-cases-en}

Table~\ref{tab:field-cases-en} presents six representative items from the domain analysis. Models more often achieve complete recall for cases involving distinctions such as birth versus baptism, alternative estimates of pandemic mortality, or global versus regional scope. By contrast, questions about fine-grained prices or event dates elicit almost exclusively partial or failed recall.

\input{tables/field_cases_en}

\FloatBarrier

\section{Conditional Perplexity as an Efficient Performance Proxy}
\label{app:ppl-probe-en}

We complement our generation-based evaluation with conditional perplexity (PPL), which scores the verified answer set directly.
For a question $q$, we place the verified answer set $\mathcal{A}$ in a short natural response and 
\begin{wrapfigure}[12]{r}{0.5\columnwidth}
\centering
\vspace{-4mm}
\includegraphics[width=\linewidth]{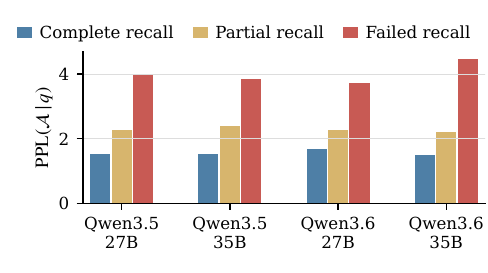}
\vspace{-8mm}
\caption{Conditional PPL grouped by response outcome. Each bar reports the geometric-mean PPL over questions graded as complete, partial, or failed recall.}
\label{fig:ppl-outcome-groups-en}
\vspace{-2mm}
\end{wrapfigure}
apply the model's native chat template with reasoning disabled. We retain the full response as autoregressive context but compute the loss only over tokens in the two answer spans:
\begin{equation}
\begin{aligned}
\mathrm{NLL}(\mathcal{A}\mid q)
&=-\frac{1}{T_{\mathcal{A}}}\sum_{t=1}^{T_{\mathcal{A}}}
\log p_{\theta}(a_t\mid q,a_{<t}),\\
\mathrm{PPL}(\mathcal{A}\mid q)
&=\exp\!\left[\mathrm{NLL}(\mathcal{A}\mid q)\right],
\end{aligned}
\label{eq:gold-span-ppl-en}
\end{equation}
where $a_{<t}$ is the preceding context. Lower PPL indicates greater conditional support for the verified answers.
Figure~\ref{fig:ppl-outcome-groups-en} shows that conditional PPL tracks answer quality across all four models. Items graded as complete recall tend to have lower PPL, whereas those graded as partial or failed recall tend to have progressively higher PPL. This outcome-conditioned view complements the PPL-binned analysis in Figure~\ref{fig:ppl-binned-outcomes-en}. Together, the two analyses support conditional PPL as a computationally efficient proxy for generation-based evaluation.

\section{Artifact Documentation and Release}
\label{app:artifact-release-en}

We will release \benchmark as an English-language evaluation dataset containing all 1,094 QA pairs, verified answer sets, and the source webpage texts. A dataset card will document its 22 fields, schema, limitations, intended use for evaluation, and the absence of train/dev/test splits. The dataset and code use the Apache License 2.0. The dataset retains source provenance. Human reviewers exclude records containing unnecessary personal identifiers, sensitive attributes, harmful allegations, or objectionable content. The release is restricted to research on factual recall and model evaluation.

\section{Human Review and Annotation Protocol}
\label{app:human-annotation-en}

Three author-annotators, all graduate researchers in computer science and artificial intelligence, conducted the final benchmark review and judge-reliability audit with institutional support. No external annotators were recruited, making recruitment and participant consent inapplicable. They retained an item only when it concerned the same subject--attribute pair, every answer was source-supported and independently verified, and the question contained no answer leakage or question--fact mismatch. They removed duplicates and unsafe content. For the reliability audit, they independently assigned complete, partial, or failed recall from the question, verified answers, and model response using the rubric in Appendix Figure~\ref{fig:closed-book-prompt-en}(b), without external tools. Appendix~\ref{app:judge-reliability-en} reports agreement with the automatic judge.

\section{Computational Resources and Software}
\label{app:compute-resources-en}

Proprietary models are evaluated through hosted APIs, while open-weight models run on an internally managed GPU cluster. Table~\ref{tab:main-results-en} reports publicly available parameter counts. Accelerator models and aggregate device hours are confidential under the infrastructure provider's policy, while direct API expenditure was approximately USD~295. Appendix~\ref{app:model-inference-en} gives the remaining inference settings.

\end{document}

%% file: tables/main_results_en.tex
\begin{wraptable}{r}{0.5\columnwidth}
\centering
\vspace{-8mm}
\captionof{table}{Evaluation results on all 1,094 questions (\%). $C/P/F$ denote complete, partial, and failed recall, and $K=C/(C+P)$. Higher $C/K$ and lower $P/F$ are better. Boldface and underlining indicate the best and second-best results within each group.}
\label{tab:main-results-en}
\small
\setlength{\tabcolsep}{0.8pt}
\renewcommand{\arraystretch}{0.92}
\begin{tabular*}{\linewidth}{@{\extracolsep{\fill}}lrrrr@{}}
\toprule
Model & $C\uparrow$ & $P\downarrow$ & $F\downarrow$ & $K\uparrow$ \\
\midrule
\rowcolor{blue!15}
\multicolumn{5}{c}{\textit{Proprietary models}} \\
\addlinespace[0.5pt]
Gemini-3.1-Pro~\citep{google2026gemini31pro} & \textbf{50.37} & \underline{47.44} & \underline{2.19} & \underline{51.50} \\
GPT-5.5~\citep{openai2026gpt55} & \underline{50.18} & \textbf{47.17} & 2.65 & \textbf{51.55} \\
Gemini-3.5-Flash~\citep{google2026gemini35flash} & 48.63 & 49.36 & \textbf{2.01} & 49.63 \\
GPT-5.6-Sol~\citep{openai2026gpt56sol} & 44.97 & 52.19 & 2.83 & 46.28 \\
Claude-Opus-4.8~\citep{anthropic2026claudeopus48} & 44.15 & 50.46 & 5.39 & 46.67 \\
Claude-Sonnet-5~\citep{anthropic2026claudesonnet5} & 37.84 & 56.12 & 6.03 & 40.27 \\
\addlinespace[0.5pt]
\midrule
\rowcolor{blue!15}
\multicolumn{5}{c}{\textit{$>100$B}} \\
\addlinespace[0.5pt]
Kimi-K3~\citep{team2026kimi} & \textbf{52.38} & \textbf{45.25} & \textbf{2.38} & \textbf{53.65} \\
Hy3~\citep{tencent2026hy3} & \underline{42.60} & 54.02 & \underline{3.38} & \underline{44.09} \\
Nemotron-3-Ultra~\citep{nvidia2026nemotron3ultra} & 38.12 & 57.95 & 3.93 & 39.68 \\
GLM-5.2~\citep{glm5team2026glm5} & 37.29 & 57.59 & 5.12 & 39.31 \\
MiniMax-M3~\citep{lai2026minimax} & 36.20 & 55.94 & 7.86 & 39.29 \\
Qwen3.5-397B~\citep{qwen2026qwen35} & 32.27 & 59.23 & 8.50 & 35.26 \\
DeepSeek-V4-Pro~\citep{xu2026deepseek} & 31.99 & 64.08 & 3.93 & 33.30 \\
DeepSeek-V4-Flash~\citep{xu2026deepseek} & 29.98 & 64.17 & 5.85 & 31.84 \\
Qwen3.5-122B~\citep{qwen2026qwen35} & 28.24 & 57.22 & 14.53 & 33.05 \\
GPT-OSS-120B~\citep{openai2025gptoss} & 22.49 & \underline{53.02} & 24.50 & 29.78 \\
\addlinespace[0.5pt]
\midrule
\rowcolor{blue!15}
\multicolumn{5}{c}{\textit{10B--100B}} \\
\addlinespace[0.5pt]
Qwen3.5-35B~\citep{qwen2026qwen35} & \textbf{21.02} & 55.94 & \underline{23.03} & \textbf{27.32} \\
Qwen3.5-27B~\citep{qwen2026qwen35} & \underline{19.47} & 55.48 & 25.05 & \underline{25.98} \\
Qwen3.6-35B~\citep{qwen2026qwen3635} & 17.00 & 59.32 & 23.67 & 22.28 \\
Seed-OSS-36B~\citep{bytedance2025seedoss} & 15.36 & 63.44 & \textbf{21.21} & 19.49 \\
OLMo-3.1-32B~\citep{olmo2025olmo3} & 14.99 & 60.69 & 24.31 & 19.81 \\
Gemma-4-26B~\citep{team2026gemma} & 13.71 & 53.84 & 32.45 & 20.30 \\
Gemma-4-31B~\citep{team2026gemma} & 11.88 & 58.59 & 29.52 & 16.86 \\
Qwen3.6-27B~\citep{qwen2026qwen36} & 10.15 & 61.79 & 28.06 & 14.10 \\
GPT-OSS-20B~\citep{openai2025gptoss} & 9.14 & \underline{49.27} & 41.59 & 15.65 \\
Gemma-4-12B~\citep{team2026gemma} & 5.85 & \textbf{42.50} & 51.65 & 12.10 \\
\addlinespace[0.5pt]
\midrule
\rowcolor{blue!15}
\multicolumn{5}{c}{\textit{$<10$B}} \\
\addlinespace[0.5pt]
Qwen3.5-9B~\citep{qwen2026qwen35} & \textbf{12.25} & 46.25 & \textbf{41.50} & \textbf{20.94} \\
Qwen3.5-4B~\citep{qwen2026qwen35} & \underline{9.69} & 39.03 & \underline{51.28} & \underline{19.89} \\
OLMo-3-7B~\citep{olmo2025olmo3} & 5.21 & 38.03 & 56.76 & 12.05 \\
Gemma-4-E4B~\citep{team2026gemma} & 2.74 & 32.82 & 64.44 & 7.71 \\
Qwen3.5-2B~\citep{qwen2026qwen35} & 1.65 & \textbf{17.00} & 81.35 & 8.82 \\
Gemma-4-E2B~\citep{team2026gemma} & 1.37 & \underline{24.31} & 74.31 & 5.34 \\
\addlinespace[0.5pt]
\bottomrule
\end{tabular*}
\vspace{-23mm}
\end{wraptable}

%% file: tables/judge_human_agreement_en.tex
\begin{center}
\begin{minipage}{\columnwidth}
\centering
\small
\setlength{\tabcolsep}{2.6pt}
\begin{tabular*}{\columnwidth}{@{\extracolsep{\fill}}lrrrr@{}}
\toprule
Model & $C_{\mathrm{human}}$ & $C_{\mathrm{judge}}$ & Agreement & Cohen's $\kappa$ \\
\midrule
Kimi-K3 & 44.46 & 45.84 & 92.77 & 0.866 \\
GPT-5.6-Sol & 36.75 & 38.76 & 92.60 & 0.859 \\
Gemini-3.1-Pro & 40.40 & 41.86 & 90.13 & 0.815 \\
Claude-Opus-4.8 & 35.21 & 36.13 & 93.36 & 0.877 \\
\midrule
Macro average & 39.21 & 40.65 & 92.22 & 0.854 \\
\bottomrule
\end{tabular*}
\captionof{table}{Agreement between the GPT-5.6-Sol judge and human annotators across four frontier models. $C$ and exact agreement are reported as percentages.}
\label{tab:judge-human-agreement-en}
\end{minipage}
\end{center}

%% file: tables/temperature_ablation_en.tex
\begin{table}[H]
\centering
\caption{Controlled temperature ablation on all 1,094 questions (\%). All settings other than temperature are held fixed. Parentheses report $T{=}0-T{=}1$ in percentage points.}
\label{tab:temperature-ablation-en}
\small
\setlength{\tabcolsep}{4pt}
\renewcommand{\arraystretch}{1.05}
\begin{tabular*}{\textwidth}{@{\extracolsep{\fill}}llllll@{}}
\toprule
Model & Temperature & $C\uparrow$ & $P\downarrow$ & $F\downarrow$ & $K\uparrow$ \\
\midrule
\multirow{2}{*}{Qwen3.6-35B-A3B~\citep{qwen2026qwen3635}}
& $T{=}1$ & 16.00 & 56.22 & 27.79 & 22.15 \\
& $T{=}0$ & 15.81~\textcolor{red!70!black}{\scriptsize($-0.18$)} & 57.50~\textcolor{red!70!black}{\scriptsize($+1.28$)} & 26.69~\textcolor{green!45!black}{\scriptsize($-1.10$)} & 21.57~\textcolor{red!70!black}{\scriptsize($-0.58$)} \\
\addlinespace[3pt]
\multirow{2}{*}{Qwen3.5-35B-A3B~\citep{qwen2026qwen35}}
& $T{=}1$ & 20.66 & 56.31 & 23.03 & 26.84 \\
& $T{=}0$ & 20.93~\textcolor{green!45!black}{\scriptsize($+0.27$)} & 56.40~\textcolor{red!70!black}{\scriptsize($+0.09$)} & 22.67~\textcolor{green!45!black}{\scriptsize($-0.37$)} & 27.07~\textcolor{green!45!black}{\scriptsize($+0.23$)} \\
\addlinespace[3pt]
\multirow{2}{*}{GPT-OSS-120B~\citep{openai2025gptoss}}
& $T{=}1$ & 22.67 & 52.93 & 24.41 & 29.99 \\
& $T{=}0$ & 22.49~\textcolor{red!70!black}{\scriptsize($-0.18$)} & 54.48~\textcolor{red!70!black}{\scriptsize($+1.55$)} & 23.03~\textcolor{green!45!black}{\scriptsize($-1.37$)} & 29.22~\textcolor{red!70!black}{\scriptsize($-0.77$)} \\
\addlinespace[3pt]
\multirow{2}{*}{Gemma-4-31B~\citep{team2026gemma}}
& $T{=}1$ & 11.79 & 58.68 & 29.52 & 16.73 \\
& $T{=}0$ & 11.33~\textcolor{red!70!black}{\scriptsize($-0.46$)} & 58.23~\textcolor{green!45!black}{\scriptsize($-0.46$)} & 30.44~\textcolor{red!70!black}{\scriptsize($+0.91$)} & 16.29~\textcolor{red!70!black}{\scriptsize($-0.44$)} \\
\bottomrule
\end{tabular*}
\end{table}

%% file: tables/exposure_regression_en.tex
\begin{table}[t]
\centering\small
\begin{tabular*}{\columnwidth}{@{\extracolsep{\fill}}lrrrr@{}}
\toprule
Side & Change & $\Delta C$ & $\Delta P$ & $\Delta F$ \\
\midrule
Majority & $+1\sigma$ & $-4.01$ & $+14.18$ & $-10.17$ \\
Minority & $+1\sigma$ & $+15.13$ & $-15.41$ & $+0.28$ \\
\bottomrule
\end{tabular*}
\caption{Estimated percentage-point changes in $C/P/F$ for a $+1\sigma$ increase in the log-transformed number of supporting documents, with the other side held fixed.}
\label{tab:exposure-regression-en}
\end{table}

%% file: tables/field_cases_en.tex
\begin{table*}[t]
\centering\footnotesize
\renewcommand{\arraystretch}{1.08}
\begin{tabular*}{\textwidth}{@{\extracolsep{\fill}}>{\raggedright\arraybackslash}p{0.14\textwidth}>{\raggedright\arraybackslash}p{0.19\textwidth}>{\raggedright\arraybackslash}p{0.14\textwidth}>{\raggedright\arraybackslash}p{0.13\textwidth}>{\raggedright\arraybackslash}p{0.28\textwidth}@{}}
\toprule
Domain & Item & Verified values & Outcome counts & Typical response pattern \\
\midrule
People, Organizations, and Events & Mother Teresa's birth date & August 26, 1910; August 27, 1910 & 18 complete; 13 partial; 1 failed & Complete answers typically identify August 26 as the date of birth and August 27 as the baptism date she regarded as her ``true birthday.'' \\
\addlinespace[2pt]
Consumer Products and Services & USD price of the Freedom 251 & USD~4; USD~7 & 0 complete; 24 partial; 8 failed & Partial answers almost uniformly repeat the widely reported price of approximately USD~4 while omitting the USD~7 report. \\
\addlinespace[2pt]
Public Health & Worldwide death toll of the 1918 Spanish Flu & 50 million; upwards of 100 million & 30 complete; 2 partial; 0 failed & Models frequently preserve both values as alternative historical estimates rather than forcing a single exact count. \\
\addlinespace[2pt]
History & Year Mary Leakey discovered the Laetoli footprints & 1975; 1978 & 0 complete; 25 partial; 7 failed & Models tend to return one familiar discovery year without recovering the less repeated account. \\
\addlinespace[2pt]
Atmospheric Science & Geographic extent of the Little Ice Age & global; regional & 23 complete; 9 partial; 0 failed & The competing scopes form a salient conceptual contrast that many models can state explicitly. \\
\addlinespace[2pt]
Applied Economics & Founding year of Grameen Bank & 1974; 1976 & 0 complete; 27 partial; 5 failed & Models generally retain one milestone year but not both reported founding dates. \\
\bottomrule
\end{tabular*}
\caption{Six representative cases illustrating domain-level variation. Outcome counts are aggregated over 32 primary model configurations as graded by GPT-5.6-Sol.}
\label{tab:field-cases-en}
\end{table*}